\documentclass[english]{article}
\usepackage[T1]{fontenc}
\usepackage[latin9]{inputenc}
\usepackage{amstext}
\usepackage{graphicx}
\usepackage{esint}

\makeatletter
\usepackage{spconf,cite}

\twoauthors
  {A. Kevin R. Moon, C. Alfred O. Hero III\sthanks{These authors were partially supported by the US National Science Foundation (NSF) under grant CCF-1217880 and a NSF Graduate Research Fellowship to KM under Grant No. F031543.}}
{University of Michigan\\
Department of EECS\\
1301 Beal Avenue, Ann Arbor, MI 48109, USA}
  {B. V{\'e}ronique Delouille\sthanks{VD acknowledges support from the Belgian Federal Science Policy Office through the ESA-PRODEX program, grant No. 4000103240.}}
{Royal Observatory of Belgium\\
SIDC\\
Avenue Circulaire 3, 1180 Uccle, Belgium}

\ninept

\makeatother

\usepackage{babel}

\begin{document}
\global\long\def\fhat#1{\hat{\mathbf{f}}_{#1,k}}

\global\long\def\Dhatl#1{\hat{\mathbf{D}}_{\phi,k(#1)}}

\title{Meta learning of bounds on the Bayes classifier error}
\maketitle
\begin{abstract}
Meta learning uses information from base learners (e.g. classifiers
or estimators) as well as information about the learning problem to
improve upon the performance of a single base learner. For example,
the Bayes error rate of a given feature space, if known, can be used
to aid in choosing a classifier, as well as in feature selection and
model selection for the base classifiers and the meta classifier.
Recent work in the field of $f$-divergence functional estimation
has led to the development of simple and rapidly converging estimators
that can be used to estimate various bounds on the Bayes error. We
estimate multiple bounds on the Bayes error using an estimator that
applies meta learning to slowly converging plug-in estimators to obtain
the parametric convergence rate. We compare the estimated bounds empirically
on simulated data and then estimate the tighter bounds on features
extracted from an image patch analysis of sunspot continuum and magnetogram
images.
\end{abstract}

\section{Introduction}

Meta learning is a method of learning from learned knowledge that
can be used to improve the performance of various learning tasks~\cite{chan1993meta,prodromidis2000meta}.
In a typical example where the learning task is classification, meta
learning is applied by first training multiple classifiers on the
training data. Each classifier may use either all of the training
data, or only a subset which may differ from other subsets in the
feature space. A test set is then fed into these classifiers and the
resulting output is then used as input to train an overall meta classifier
such as a majority vote or weighted majority vote. Other variations
on meta learning applied to classification exist~\cite{prodromidis2000meta,chan1993metaArbiter,chan1993metaCombine}.

Meta learning can incorporate information about the feature space
that is independent of the classifiers such as the Bayes error rate
(BER). Consider the problem of classifying a feature vector $x$ into
one of two classes $C_{1}$ or $C_{2}$. Denote the \emph{a priori
}class probabilities as $q_{1}=\Pr(C_{1})>0$ and $q_{2}=\Pr(C_{2})=1-q_{1}>0$.
The conditional densities of $x$ given that $x$ belongs to $C_{1}$
or $C_{2}$ are denoted by $f_{1}(x)$ and $f_{2}(x)$, respectively,
and the Bayes classifier assigns $x$ to $C_{1}$ if and only if $q_{1}f_{1}(x)>q_{2}f_{2}(x)$.
If $p(x)=q_{1}f_{1}(x)+q_{2}f_{2}(x)$, the average error rate of
this classifier, known as the BER, is \begin{eqnarray}
P_{e}^{*} & = & \int\min\left(\Pr\left(C_{1}|x\right),\Pr\left(C_{2}|x\right)\right)p(x)dx\nonumber \\
 & = & \int\min\left(q_{1}f_{1}(x),q_{2}f_{2}(x)\right)dx.\label{eq:BER}\end{eqnarray}

The BER is the minimum classification error rate that can be achieved
by any classifier on $x$'s feature space~\cite{hastie2009learning}.
Because of this property, the BER can be used in a meta learning problem
where the base classifiers are trained on different feature spaces
by weighting the output of the base classifiers based on the Bayes
error of the underlying feature space. If a given feature space results
in a lower Bayes error than another feature space, then the output
of the corresponding classifier would have a higher weight as it would
presumably perform better than a classifier on the alternate feature
space.

The BER can be used at other stages of meta learning such as in the
selection of the base classifiers and model selection. This is because
the BER provides a benchmark for classification on a given feature
space. If a specific classifier applied to the feature space yields
an estimated error rate that is significantly above the BER, then
it is likely that a different classifier or parameters may result
in a lower error rate. On the other hand, if the classifier's estimated
error rate is below the BER, then the classifier is likely to be overfitting
the data and may not generalize well to new samples from the feature
space. A different classifier or parameters may then be chosen. This
technique can also be applied in the traditional supervised learning
approach where a single classifier is used.

The BER can also be beneficial for feature selection in classification
problems. The BER is monotonic in the number of features in the sense
that increasing the number of features does not decrease the accuracy
of the Bayes classifier. However, for many classifiers, including
irrelevant features can decrease the prediction accuracy~\cite{aha1991instance,kohavi1997wrappers}.
Including a large number of features can also be computationally burdensome
and create difficulties in storage and memory~\cite{aha1992feature,liu2007feature}.
Thus from a practical perspective, using only a subset of the features
may result in better performance. If the BER is known for all subsets
of features, then a logical method of feature selection would be to
choose the smallest subset of features such that the BER of that subset
is negligibly larger than the BER of the full feature space~\cite{carneiro2005minimum}.
The eliminated features could be considered redundant or irrelevant
since including them in the classification leads to a neglible improvement
in accuracy.

Unfortunately, computing the BER requires perfect knowledge of the
underlying data distributions, which is rarely available. Even for
parametric models of the densities, Eq.~\ref{eq:BER} requires multi-dimensional
integration and has no closed form solution for many models. Evaluating
the BER in these cases involves computationally intensive numerical
integration, especially for high dimensions. For these reasons, many
feature selection algorithms have focused on other optimality criteria
such as minimizing the prediction error of a specific classifier~\cite{kohavi1997wrappers}
or maximizing the statistical dependency between the feature subset
and class assignments via some criterion such as mutual information
or correlation~\cite{peng2005feature}. However, selecting features
by minimizing the classifier prediction error can be computationally
intensive and only provides a solution for the specified classifier.
Additionally, other methods based on maximizing statistical dependency
can be too restrictive or otherwise problematic~\cite{frenay2012feature}.

Given these problems, many bounds on the BER have been derived that
are related to $f$-divergences~\cite{chernoff1952measure,berisha2014bound,avi1996bound,hashlamoun1994bound}.
These bounds have been used in applications involving the BER including
feature selection~\cite{berisha2014bound,xuan2006feature,zhang2007gene,bruzzone1995feature,guorong1996feature}.

Accurate estimation of these bounds on the BER requires accurate estimation
of an $f$-divergence functional, often in a nonparametric setting.
Until recently, little has been known about the properties of nonparametric
$f$-divergence estimators such as convergence rates and the asymptotic
distribution. In Moon and Hero~\cite{moon2014isit}, it was shown
that the bias of simple density plug-in estimators of $f$-divergence
converges very slowly to zero when the dimension of the feature space
is high, which limits their utility. Nguyen et al~\cite{nguyen2010div}
proposed an $f$-divergence estimation method based on estimating
the likelihood ratio of two densities that achieves the parametric
mean squared error (MSE) convergence rate when the densities are sufficiently
smooth. However, this method can be computationally intensive for
large sample sizes and the asymptotic distribution of the estimator
is currently unknown. Berisha et al~\cite{berisha2014bound} also
proposed a consistent estimator of specific bounds on the BER based
on the construction of a minimal spanning tree (MST) that does not
require density estimation. However, the convergence rate of this
estimator is unknown and it is restricted to specific BER bounds instead
of $f$-divergences in general. Finally, other $f$-divergence estimators
have been proposed that achieve the parametric  rate when the densities
are sufficiently smooth~\cite{krishnamurthy2014divergence,singh2014exponential,singh2014renyi}.
However, some of these estimators are restricted to certain subsets
of $f$-divergences, and they require an optimal kernel which can
be difficult to implement and compute.

Many of these problems can be countered effectively by using meta
learning. While meta learning was described above in the classification
setting, it can also be applied to estimation to improve the convergence
rates. This is typically done by taking a weighted sum of base estimators
that individually converge slowly. Then by an appropriate choice of
weights, the weighted ensemble estimator converges rapidly to the
true value. For example, Sricharan et al~\cite{sricharan2013ensemble}
derived a nonparametric estimator of generalized entropy functionals
that converges at the parametric rate by using simple plug-in density
estimators as the base estimators. More recently, similar theory was
applied by Moon and Hero~\cite{moon2014isit,moon2014nips} to obtain
a nonparametric $f$-divergence functional estimator based on a weighted
ensemble of $k$-nearest neighbor (nn) estimators. This estimator
enjoys the advantages of being simple to implement and achieving the
parametric convergence rate when the densities are sufficiently smooth. 

In this paper, we focus on estimating multiple bounds on the Bayes
error derived from $f$-divergence functionals in a nonparametric
setting using the weighted $k$-nn estimator from~\cite{moon2014isit,moon2014nips}.
We first estimate the bounds on simulated data where the true BER
is computable. This gives a guide for the empirical utility of each
bound. We then apply this to real data by estimating the bounds on
the BER for the classification of sunspot images using the features
derived in~\cite{moon2015partII}. This gives a measure of the utility
of the derived feature space in this supervised setting. We also compare
the results to those obtained usinge the MST estimator~\cite{berisha2014bound}.
The paper is outlined as follows. Section~\ref{sec:Estimation} describes
the weighted $k$-nn estimator of $f$-divergence functionals while
Section~\ref{sec:Bounds} provides the bounds on the Bayes error
and their relation to $f$-divergences. In Section~\ref{sec:simulations},
the simulated results are presented. Section~\ref{sec:sunspot} describes
the sunspot data and presents the estimated bounds on the BER. Section~\ref{sec:conclusion}
concludes.

\section{Meta Learning of $f$-Divergence Functionals}

\label{sec:Estimation}If $f_{1}$ and $f_{2}$ are $d$-dimensional
probability densities with common support, then the $f$-divergence
between $f_{1}$ and $f_{2}$ has the following form~\cite{csiszar1967div}:\begin{equation}
D_{\phi}(f_{1},f_{2})=\int\phi\left(\frac{f_{1}(x)}{f_{2}(x)}\right)f_{2}(x)dx.\label{eq:fDiv}\end{equation}
For $D_{\phi}$ to be considered a true divergence, the function $\phi$
must be convex and $\phi(1)=0$. This ensures that $D_{\phi}$ is
nonnegative and that $D_{\phi}(f_{1},f_{2})=0$ if and only if $f_{1}=f_{2}$
which is the definition of divergence. As for general divergences,
$f$-divergences are not required to be symmetric or satisfy the triangle
inequality.

In this work, we are concerned with a broader class of functions that
we call $f$-divergence functionals. This class consists of functions
of the form in Eq.~\ref{eq:fDiv} except that we do not require $\phi$
to be convex or that $\phi(1)=0$. Working with $f$-divergence functionals
instead of only $f$-divergences provides greater flexibility in bounding
the BER.

Assume that the densities $f_{1}$ and $f_{2}$ have a common bounded
support set $\mathcal{S}$; $f_{1}$ and $f_{2}$ are strictly lower
bounded; and $f_{1}$, $f_{2}$, and $\phi$ are smooth. Assume that
$T=N+M$ i.i.d. realizations $\mathcal{X}_{T}=\left\{ \mathbf{X}_{1},\dots,\mathbf{X}_{N},\mathbf{X}_{N+1},\dots,\mathbf{X}_{N+M}\right\} $
are available from the density $f_{2}$ and $M$ i.i.d. realizations
$\mathcal{Y}_{M}=\left\{ \mathbf{Y}_{1},\dots,\mathbf{Y}_{M}\right\} $
are available from the density $f_{1}$, where $M$ is proportional
to $T$. Under these assumptions, there exists a nonparametric estimator
of $D_{\phi}$ that achieves the parametric MSE rate of $O\left(\frac{1}{T}\right)$.
This estimator first calculates an ensemble of $k$-nn density estimators
of the densities $f_{1}$ and $f_{2}$ at the points $\left\{ \mathbf{X}_{1},\dots,\mathbf{X}_{N}\right\} $
using different values of $k$. Then for each $k$, a base plug-in
estimator of $D_{\phi}$ is calculated by taking the empirical average
of the $\phi$ evaluated at the likelihood ratio of the estimated
densities. From~\cite{moon2014isit}, the bias and variance of these
base estimators is known. Then using the theory of optimally weighted
ensemble estimation~\cite{sricharan2013ensemble}, an estimator with
low bias can be obtained by taking a weighted sum of the base estimators
using the appropriate weights. Details are given in the following.

Let $k\leq M$ and let $\mathbf{\rho}_{2,k}(i)$ be the distance of
the $k$th nearest neighbor of $\mathbf{X}_{i}$ in $\left\{ \mathbf{X}_{M+1},\dots,\mathbf{X}_{N}\right\} $.
Similarly, define $\mathbf{\rho}_{1,k}(i)$ be the distance of the
$k$th nearest neighbor of $\mathbf{X}_{i}$ in $\left\{ \mathbf{Y}_{1},\dots,\mathbf{Y}_{M}\right\} $.
Then the $k$-nn density estimates at the point $\mathbf{X}_{i}$
are~\cite{loftsgaarden1965knn}\[
\fhat j(\mathbf{X}_{i})=\frac{k}{M\bar{c}\mathbf{\rho}_{j,k}^{d}(i)},\]
where $\bar{c}$ is the volume of a $d$-dimensional unit ball. The
functional $D_{\phi}$ is then approximated as \[
\hat{\mathbf{D}}_{\phi,k}=\frac{1}{N}\sum_{i=1}^{N}\phi\left(\frac{\fhat 1(\mathbf{X}_{i})}{\fhat 2(\mathbf{X}_{i})}\right).\]

Now choose an ensemble of positive real numbers $\bar{\ell}=\left\{ \ell_{1},\dots,\ell_{L}\right\} $
where $L>d-1$ and let $k(\ell)=\ell\sqrt{M}$. It was shown in~\cite{moon2014isit}
that the bias and variance of $\Dhatl{\ell}$ are\begin{eqnarray*}
\text{Bias}\left(\Dhatl{\ell}\right) & = & \sum_{j=1}^{d}O\left(\left(\frac{\ell}{\sqrt{M}}\right)^{\frac{j}{d}}\right)+O\left(\frac{1}{\sqrt{M}}\right),\\
\text{Var}\left(\Dhatl{\ell}\right) & = & O\left(\frac{1}{N}+\frac{1}{M}\right).\end{eqnarray*}

Let $w$ be a vector of weights with length $L$ and define $\hat{\mathbf{D}}_{\phi,w}:=\sum_{\ell\in\bar{\ell}}w(\ell)\Dhatl{\ell}$.
From the theory of optimally weighted ensemble estimation~\cite{sricharan2013ensemble},
there exists a weight vector $w_{0}$ such that the MSE of $\hat{\mathbf{D}}_{\phi,w_{0}}$
is $O\left(\frac{1}{T}\right)$. The weight vector $w_{0}$ achieves
this by essentially zeroing out the lower order bias terms at the
expense of a slight increase in the variance. $w_{0}$ can be found
via an offline convex optimization problem that only depends on the
sample size $T$ and the basis functions $\ell^{\frac{j}{d}}$. See~\cite{moon2014isit,sricharan2013ensemble,moon2014nips}
for more details.

\section{Bounds on the Bayes Error Rate}

\label{sec:Bounds}Multiple upper and lower bounds on the BER related
to $f$-divergences exist. A classical bound is the Chernoff bound~\cite{chernoff1952measure}.
It is derived from the fact that for $a,\, b>0$, $\min(a,b)\leq a^{\alpha}b^{1-\alpha}$
$\forall\alpha\in(0,1)$. Replacing the minimum function in Eq.~\ref{eq:BER}
with this bound gives \begin{equation}
P_{e}^{*}\leq q_{1}^{\alpha}q_{2}^{1-\alpha}c_{\alpha}(f_{1},f_{2}),\label{eq:chernoff}\end{equation}
where $c_{\alpha}(f_{1},f_{2})=\int f_{1}^{\alpha}(x)f_{2}^{1-\alpha}(x)dx$
is the Chernoff $\alpha$-coefficient. The Chernoff coefficient is
found by minimizing the right hand side of Eq.~\ref{eq:chernoff}
with respect to $\alpha$:\begin{equation}
c^{*}(f_{1},f_{2})=\min_{\alpha\in(0,1)}\int f_{1}^{\alpha}(x)f_{2}^{1-\alpha}(x)dx.\label{eq:chernoffCoeff}\end{equation}
Combining this with Eq.~\ref{eq:chernoff} gives an upper bound on
the BER.

In general, the Chernoff bound is not very tight. A tighter bound
was presented in~\cite{berisha2014bound}. Consider the following
quantity:\begin{eqnarray}
\tilde{D}_{q_{1}}(f_{1},f_{2}) & = & 1-4q_{1}q_{2}\int\frac{f_{1}(x)f_{2}(x)}{q_{1}f_{1}(x)+q_{2}f_{2}(x)}dx\label{eq:Dpgood}\\
 & = & \int\frac{\left(q_{1}f_{1}(x)-q_{2}f_{2}(x)\right)^{2}}{q_{1}f_{1}(x)+q_{2}f_{2}(x)}dx.\label{eq:Dpbad}\end{eqnarray}
It was shown in~\cite{berisha2014bound} that the BER $P_{e}^{*}$
is bounded above and below as follows:\[
\frac{1}{2}-\frac{1}{2}\sqrt{\tilde{D}_{q_{1}}(f_{1},f_{2})}\leq P_{e}^{*}\leq\frac{1}{2}-\frac{1}{2}\tilde{D}_{q_{1}}(f_{1},f_{2}).\]

Arbitrarily tight upper and lower bounds to the BER were given in~\cite{avi1996bound}.
We consider only the lower bound here. Define\[
g_{\alpha}(f_{1},f_{2})=\ln\left(\frac{1+e^{-\alpha}}{\exp\left(\frac{-\alpha q_{1}f_{1}(x)}{p(x)}\right)+\exp\left(\frac{-\alpha q_{2}f_{2}(x)}{p(x)}\right)}\right),\]
where $p(x)=q_{1}f_{1}(x)+q_{2}f_{2}(x)$ as before and $\alpha>0$.
Then the BER is bounded below as\begin{equation}
P_{e}^{*}\geq\frac{1}{\alpha}\int g_{\alpha}(f_{1},f_{2})p(x)dx=:G_{\alpha}(f_{1},f_{2}).\label{eq:LowerBound}\end{equation}

The functionals in Eqs.~\ref{eq:chernoff} and \ref{eq:Dpgood}-\ref{eq:LowerBound}
all contain the form in Eq.~\ref{eq:fDiv}. To see this, note that
for the Chernoff $\alpha$ coefficient, $\phi(t)=t^{\alpha}$. For
the $\tilde{D}_{q_{1}}$ based bounds, the functions are more complicated
with $\phi(t)=\frac{4q_{1}q_{2}t}{q_{2}+q_{1}t}$ and $\phi(t)=\frac{\left(q_{1}t-q_{2}\right)^{2}}{q_{1}t+q_{2}}$
for Eqs.~\ref{eq:Dpgood} and~\ref{eq:Dpbad}, respectively. The
functions are even more complex for Eq.~\ref{eq:LowerBound}. However,
if $t=\frac{f_{1}(x)}{f_{2}(x)}$, then \begin{eqnarray*}
\exp\left(\frac{-\alpha q_{1}f_{1}(x)}{p(x)}\right) & = & \exp\left(\frac{-\alpha q_{1}}{q_{1}+q_{2}t^{-1}}\right),\\
\exp\left(\frac{-\alpha q_{2}f_{2}(x)}{p(x)}\right) & = & \exp\left(\frac{-\alpha q_{2}}{q_{2}+q_{1}t}\right).\end{eqnarray*}
Substituting these expressions into $G_{\alpha}(f_{1},f_{2})$ gives
the required form. Thus we can use the optimally weighted ensemble
divergence estimator from Section~\ref{sec:Estimation} to estimate
all of these bounds on the Bayes error. To estimate $c^{*}(f_{1},f_{2})$,
we estimate $c_{\alpha}(f_{1},f_{2})$ for multiple values of $\alpha$
(e.g. $0.01,\,0.02,\dots,0.99$) and choose the minimum.

\section{Simulations}

\label{sec:simulations}In addition to the weighted $k$-nn estimator,
we use an alternate estimator for $\tilde{D}_{q_{1}}$ based on an
extension of the Friedman-Rafsky (FR) multivariate two sample test
statistic for comparison~\cite{friedman1979test}. This estimator
is derived from the MST of the combined data set $\mathcal{X}_{T}\cup\mathcal{Y}_{M}$
and does not require direct estimation of the densities $f_{1}$ and
$f_{2}$~\cite{berisha2014bound,berisha2014empirical}. However,
the convergence rate and asymptotic distribution of this estimator
are currently unknown.

\begin{figure}[!h]
\centering

\includegraphics[width=1\columnwidth]{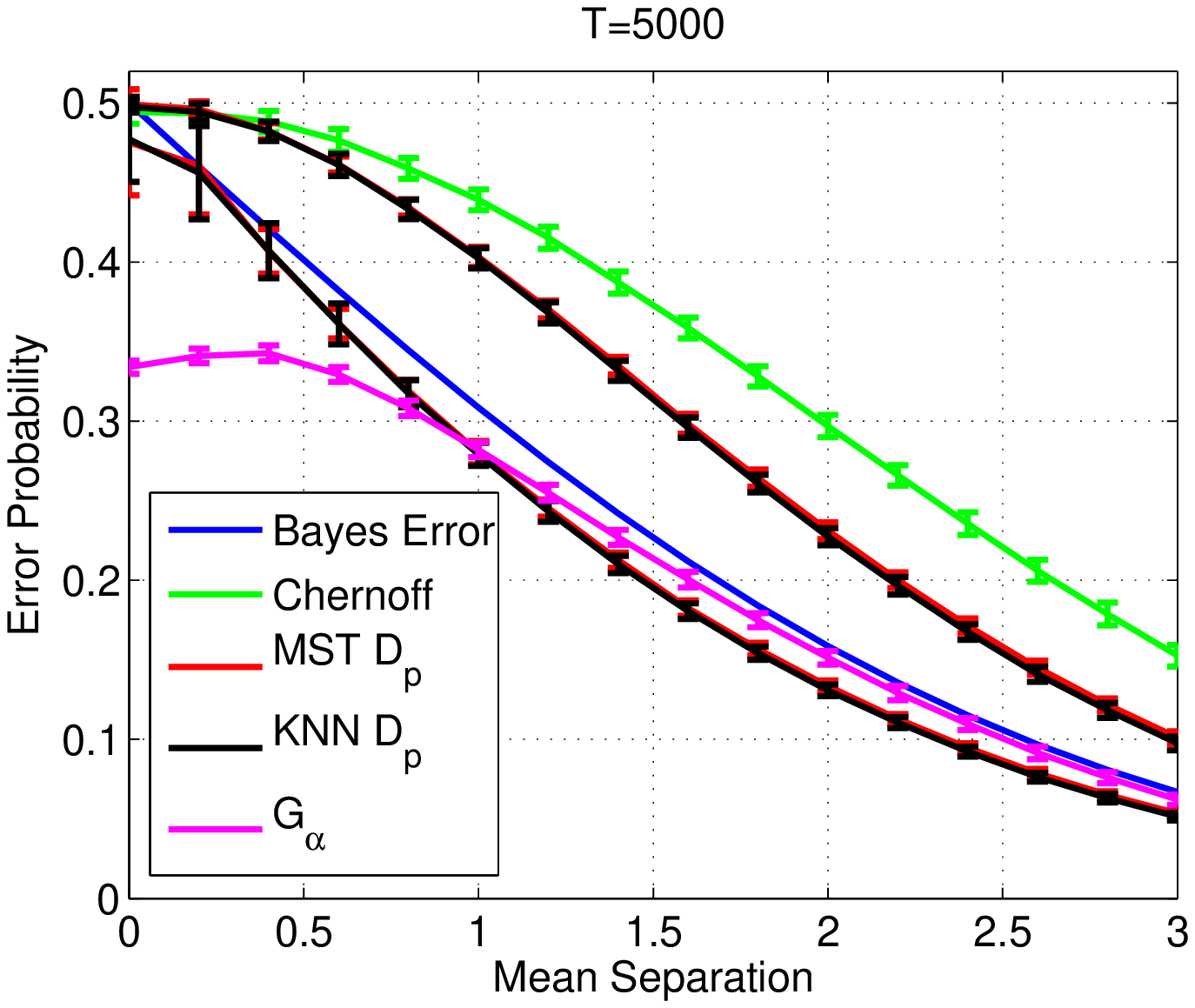}

\includegraphics[width=1\columnwidth]{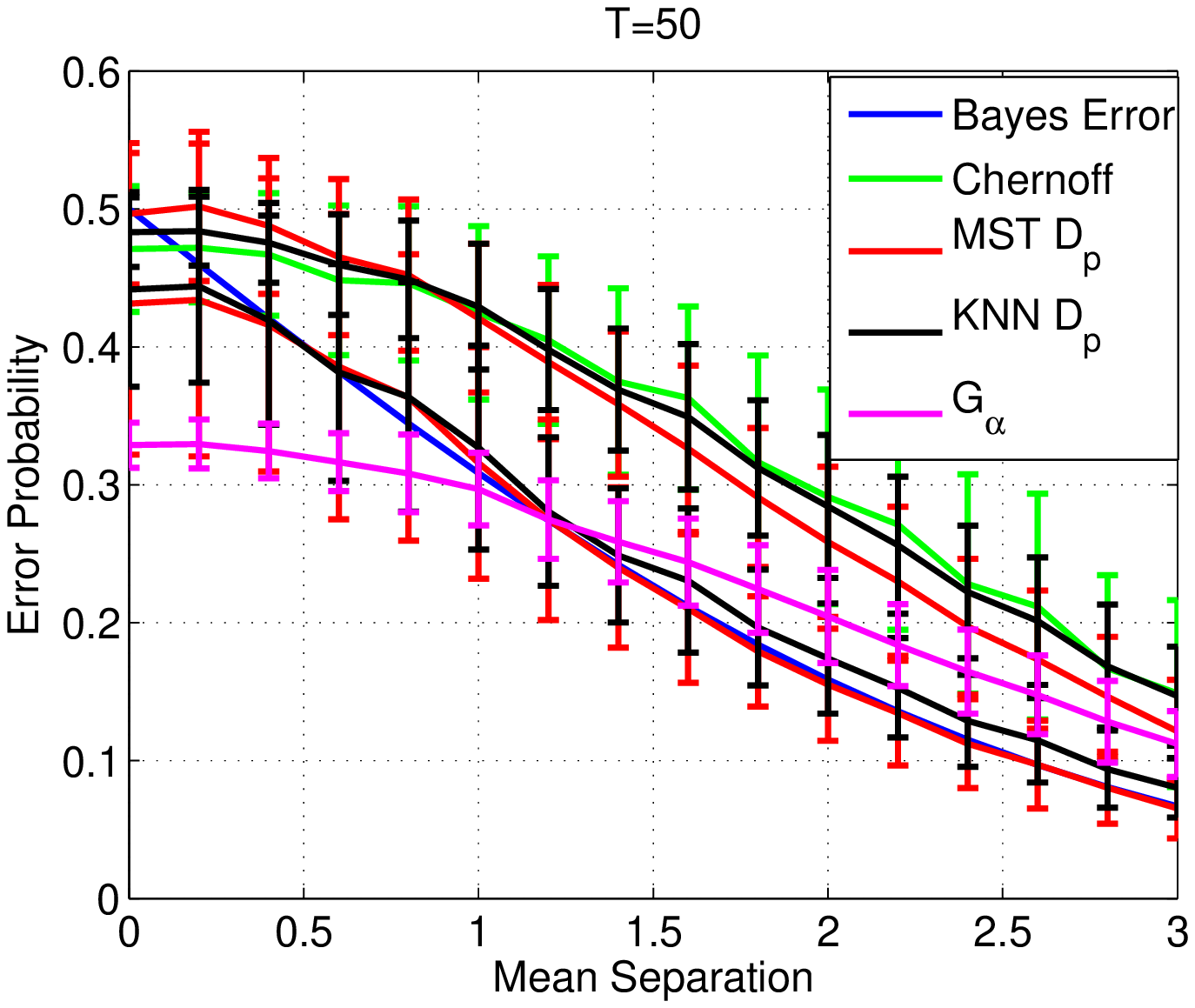}

\caption{Estimated bounds on the Bayes error rate for two unit variance Gaussians
with dimension $d=5$, varying sample sizes ($T=5000,\,50$), and
varying means over 200 trials. Error bars correspond to a single standard
deviation. The $\tilde{D}_{q_{1}}$ based lower bounds are close to
the actual Bayes error for both the large and small sample regimes
but are much more variant with a smaller sample size. The arbitrarily
tight lower bound ($G_{\alpha}$ with $\alpha=500$) is very close
to the Bayes error when $T=5000$ and when the Bayes error is low.
\label{fig:GaussSamples}}
\end{figure}

To compare the estimation performance of the various bounds on the
BER, we consider 200 trials of two samples from two Gaussian distributions
with unit variance and varying mean. In practice, we use a leave one
out approach for the weighted $k$-nn estimator and so the number
of samples from both distributions is equal to $T$. In the first
experiment, we fix the dimension $d=5$ and vary the number of samples
from each distribution. Figure~\ref{fig:GaussSamples} shows the
cases where $T=5000$ and $50$. We choose $\alpha=500$ for $G_{\alpha}$.
In the large sample regime, the bounds vary smoothly as the separation
between the means of the distributions increases. The two methods
for estimating $\tilde{D}_{q_{1}}$ have nearly identical results
when Eq.~\ref{eq:Dpgood} is used for the weighted $k$-nn method.
If Eq.~\ref{eq:Dpbad} is used, then the estimated bounds (not shown)
are inaccurate. This underscores the importance of using an appropriate
representation of the function $\phi$ when using plug-in based estimation
methods as numerical errors may lead to varying results.

In the low sample regime, the estimates have much higher variance
and are more biased as the lower bounds often cross the Bayes error.
However, the $\tilde{D}_{q_{1}}$ based lower bounds are still fairly
close to the true BER and are thus valuable for assessing the potential
performance of a given feature space. Increasing the sample size to
as little as 150 greatly improves the performance (not shown).

\begin{figure}[!h]
\centering

\includegraphics[width=1\columnwidth]{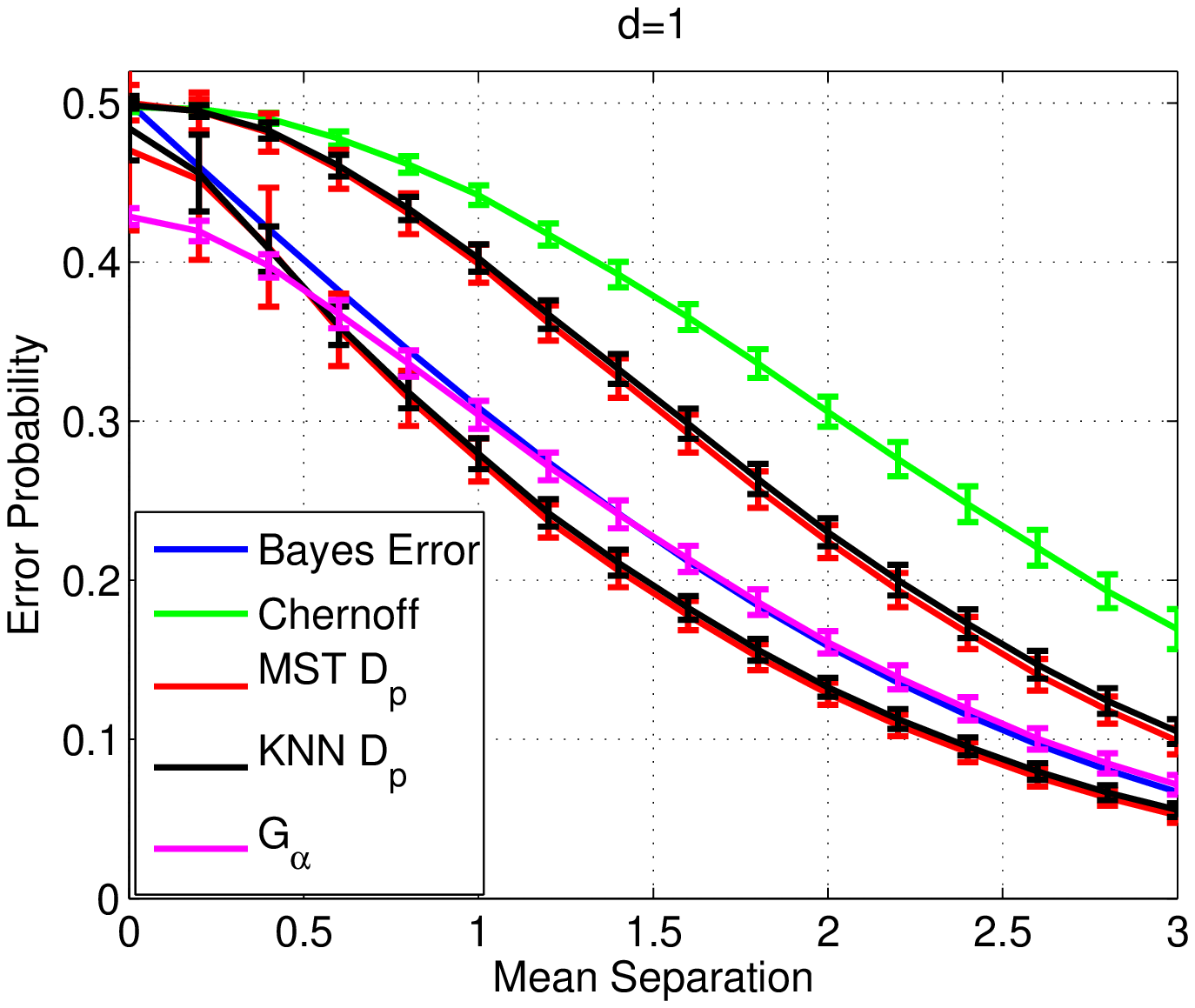}

\includegraphics[width=1\columnwidth]{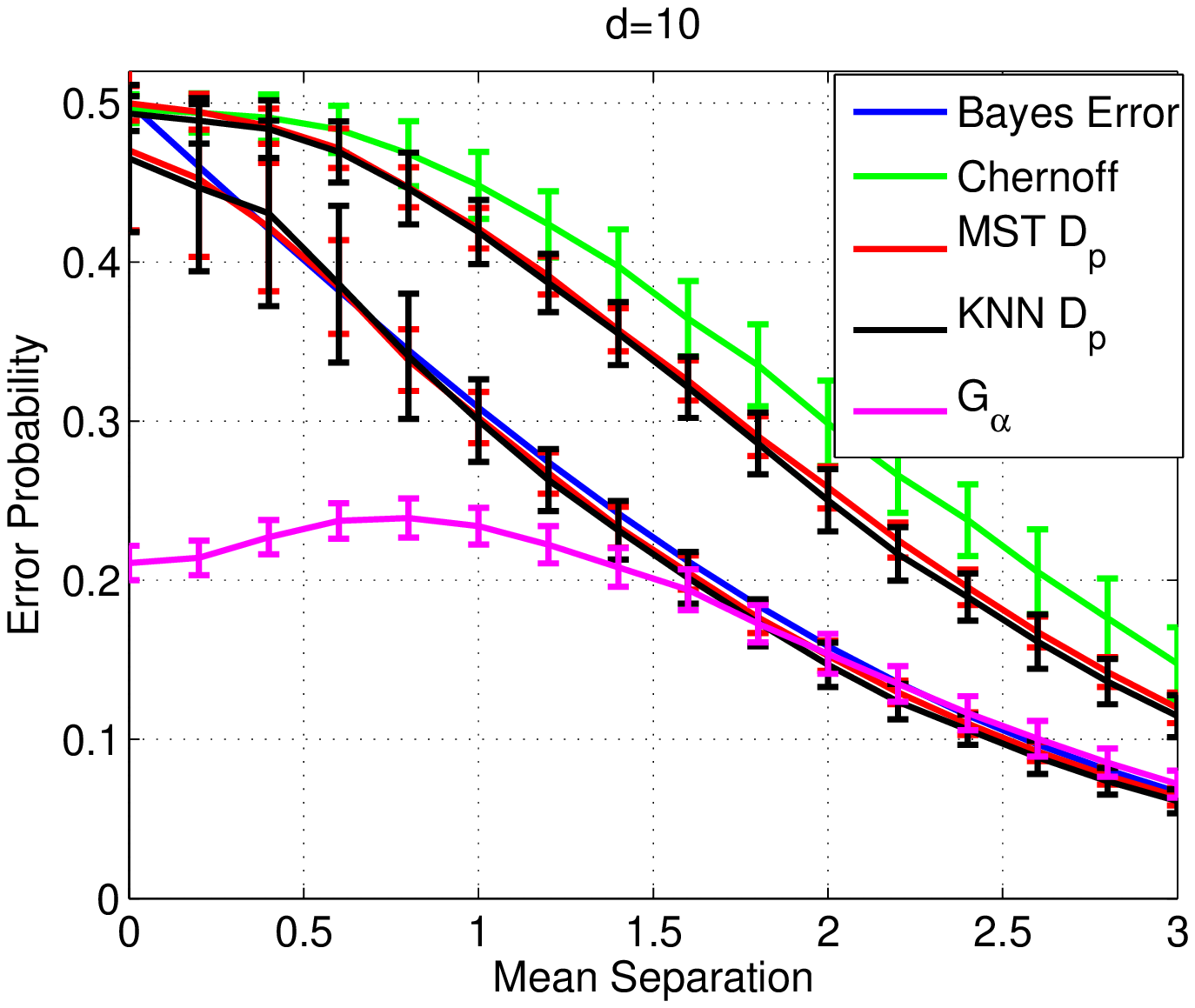}

\caption{Estimated bounds on the Bayes error rate for two unit variance Gaussians
with varying dimension ($d=1,\,10$) and a fixed sample size of $T=1000$
over 200 trials. The estimated $\tilde{D}_{q_{1}}$ based bounds are
more biased and variant when the dimension is higher. \label{fig:GaussDim}}
\end{figure}

In the second experiment, we fixed the number of samples at $T=1000$
and varied the dimension. The results for $d=1$ and $10$ are given
in Fig.~\ref{fig:GaussDim}. In the higher dimension, the $\tilde{D}_{q_{1}}$
lower bounds are closer to the BER which results in these estimates
crossing over the BER more often. The variance in all of the estimates
is also higher when $d=10$.

Several trends are apparent in both Figs.~\ref{fig:GaussSamples}
and~\ref{fig:GaussDim}. One is that the variance of the $\tilde{D}_{q_{1}}$
lower bounds decreases as the BER decreases. In general, the MST-based
estimator is more variant than the $k$-nn estimator except when the
dimension or number of samples is high (e.g. $d=10$ or $T=5000$).
This is not a substantial problem as an accurate estimate of the BER
is less useful at higher values. This is because if the BER is around
0.4, then the feature space being considered does not improve the
classification much beyond random guessing. Thus time and energy may
be better spent on finding a new feature space for the problem instead
of attempting to achieve the BER on the given feature space.

Another observation is that for $d>1$, the $G_{\alpha}$ based lower
bound is not tight for higher BER when using $\alpha=500$. Increasing
$\alpha$ does not substantially improve the tightness at these values
due to numerical precision errors. However, it may be possible to
manipulate the expression for $g_{\alpha}$ so that this is not an
issue.

Overall, these results suggest that estimating the $\tilde{D}_{q_{1}}$
lower bound provides a value that is fairly close to the true BER.
The weighted $k$-nn estimator appears to be less variant than the
MST based estimator except when the dimension or number of samples
is sufficiently high. Thus we recommend using the $\tilde{D}_{q_{1}}$
bounds to estimate the location of the BER. If this gives a range
for the BER that is low (approximately less than 0.2) and there are
enough samples, then $G_{\alpha}$ may be estimated for a more precise
estimate of the BER. Similar results are obtained for truncated Gaussians.

\section{Bounding the Bayes Error of Sunspot Images}

\label{sec:sunspot}We estimate bounds on the BER of a sunspot image
classification problem. Sunspots (SS) are dark areas seen in white
light images of the Sun. They correspond to regions of locally enhanced
magnetic field, as can be seen on magnetogram. SS groups are commonly
classified using the Mount Wilson classification scheme, which categorizes
them by eye based on their morphological features in continuum (white
light intensity) and magnetogram (magnetic field value) images. Several
studies have shown that major solar eruptive events are strongly correlated
with complex SS groups (designated as $\beta\gamma$ or $\beta\gamma\delta$
groups) and less so with simple SSs ($\alpha$ or $\beta$ groups)~\cite{warwick1966sunspot,sammis2000sunspot}. 

Recent work has focused on clustering SSs using an image patch analysis
of continuum and magnetogram images and by applying dictionary learning
on the collection of patches~\cite{moon2015partII,moon2014icip}.
Two main approaches were used in~\cite{moon2015partII}. In the first
approach, a dictionary is learned for each SS image pair. The pairwise
difference between these dictionaries is calculated by comparing the
subspaces spanned by the dictionaries using the Grassmannian projection
metric. These pairwise distances are then fed into a clustering algorithm.
For the second approach, a single dictionary is learned from the combined
collection of image patches from all SS image pairs. The dictionary
coefficients corresponding to a single SS image pair are treated as
samples from a distribution. The pairwise distances between these
collections of coefficient samples is calculated by estimating the
Hellinger distance of the underlying distribution and these distances
are then fed into a clustering algorithm.

The resulting clusterings from these two approaches were found to
be correlated somewhat with the Mount Wilson classification scheme.
In this work, we estimate the ability of the associated feature spaces
of these two approaches to classify a SS as \lq complex' or \lq
simple' by estimating bounds on the Bayes error. We do this by estimating
both the lower and upper bounds formed from $\tilde{D}_{q_{1}}$ using
both the weighted $k$-nn and MST estimators for the Grassmannian
approach from the pairwise distances. Bootstrapping is used on the
weighted $k$-nn estimators to calculate confidence intervals. For
the Hellinger distances, we only use the MST estimator as the $k$-nn
density estimator is not easily defined in the space of probability
distributions. 

We use the same image pairs as in~\cite{moon2015partII} except we
exclude the $\alpha$ groups. This is to keep the number of simple
and complex image pairs roughly the same (192 and 182, respectively).
As in~\cite{moon2015partII}, we consider two types of areas: the
area within the sunspot and the area near the corresponding neutral
line as determined from magnetogram images. The morphology of both
of these areas are taken into account in the Mount Wilson classification.
The two metrics, Grassmannian and Hellinger distance, are applied
within these areas separately and a weighted average is taken of the
two distances. For example, if $D_{G,n}$ is the distance matrix comparing
the dictionaries learned from each SS's neutral line using the Grassmannian
metric, and if $D_{G,s}$ is the distance matrix comparing the dictionaries
learned from within the sunspots, then define $D_{G}(r)=rD_{G,n}+(1-r)D_{G,s}$
with $0\leq r\leq1$. The distance matrix $D_{G}(r)$ is then used
to estimate the bounds on the Bayes error for a variety of weights.
For comparison, we calculate the error rate of a support vector machine
(SVM) classifier with a Gaussian kernel using 10-fold cross validation
to select the parameters.

\begin{figure}[!t]
\centering

\includegraphics[width=1\columnwidth]{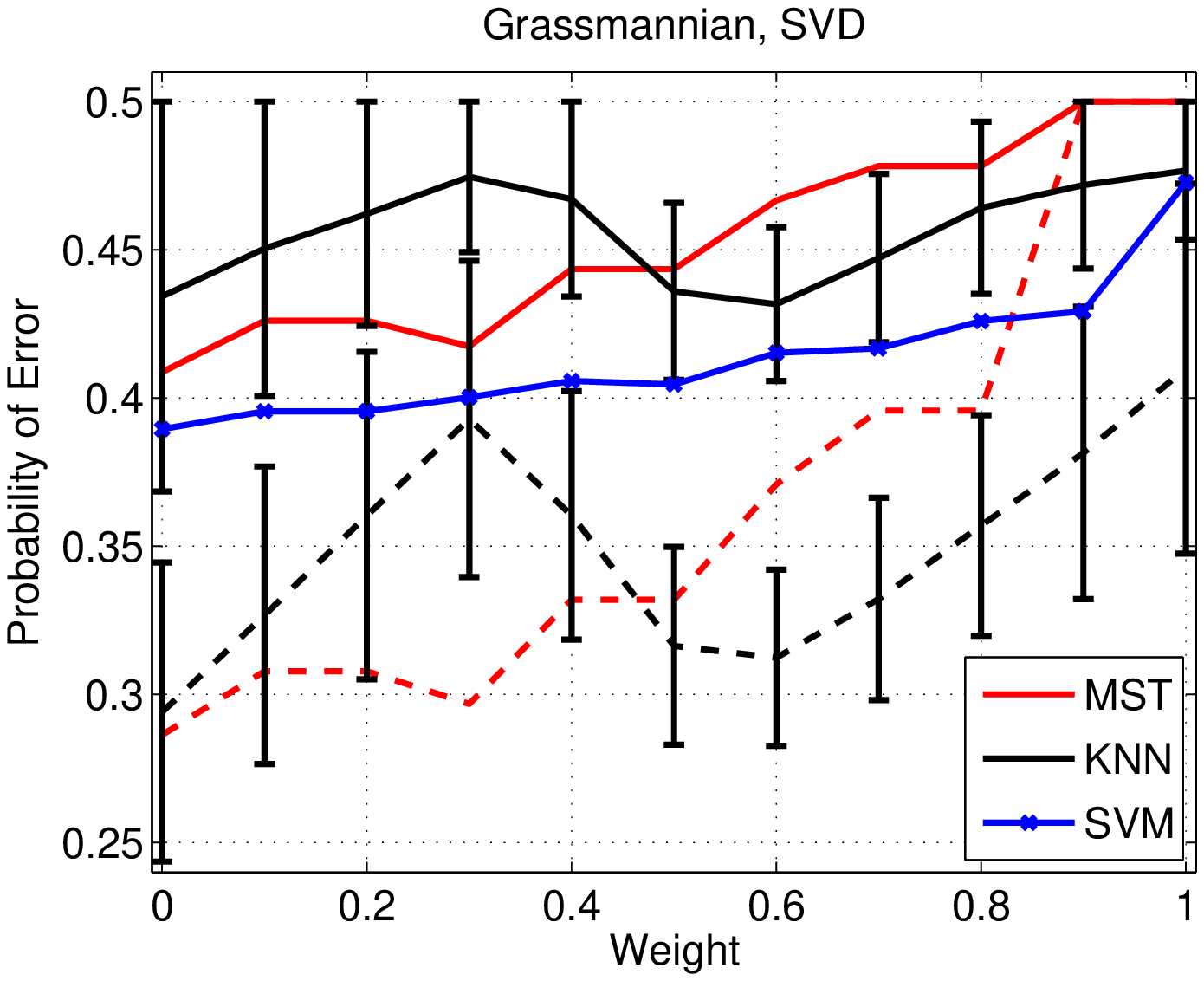}

\includegraphics[width=1\columnwidth]{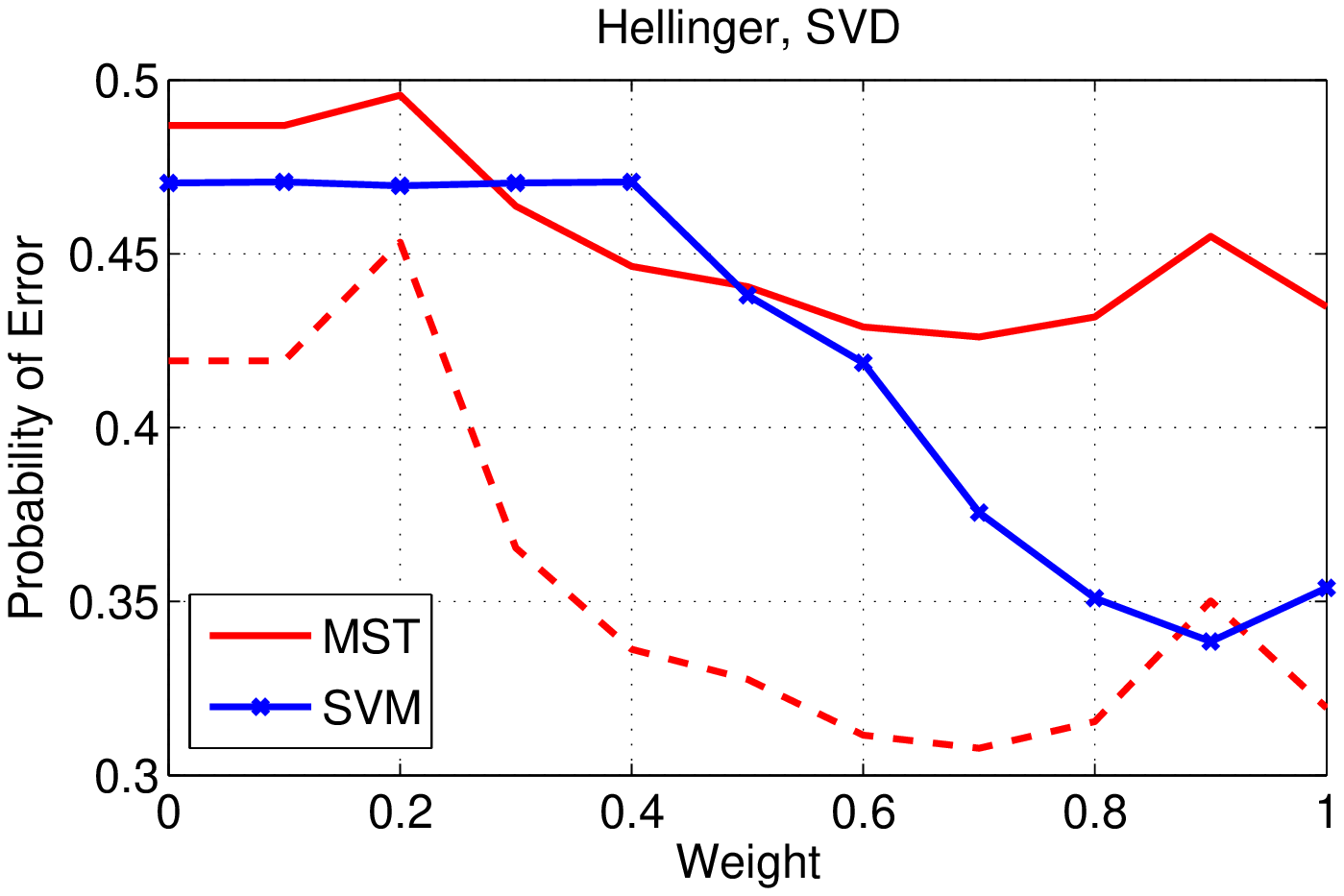}

\caption{$\tilde{D}_{q_{1}}$-based upper (plain line) and lower (dashed line)
bounds on the Bayes error when classifying sunspot groups as simple
or complex for a variety of weights compared to the error from an
SVM classifier  using SVD dictionaries. A weight of $r=0$ corresponds
to using only the data from within the sunspots while $r=1$ corresponds
to using only the neutral line data. Confidence intervals on the weighted
$k$-nn estimators are calculated via bootstrapping. The area around
the neutral line and sunspots give better results when using the Hellinger
and Grassmannian metrics, respectively. \label{fig:sunspot}}

\end{figure}

Two dictionary learning methods are used: the singular value decomposition
(SVD) and nonnegative matrix factorization (NMF). Figure~\ref{fig:sunspot}
shows the estimated bounds when using SVD. Several patterns are apparent
in the results. Both the estimated bounds and the SVM error rate generally
increase as the weight $r$ increases when the Grassmannian metric
on individual dictionaries is used. This indicates that the dictionaries
extracted from within the sunspots are more relevant to this classification
problem than the dictionaries from the neutral line. The opposite
occurs when the Hellinger distance is used on the dictionary coefficients.
In this case, the estimated bounds and SVM error rate are generally
lower when the weight $r$ favors the neutral line data. Strong spatial
gradients in the magnetogram along the neutral line are often associated
with complex SSs. Since the learned dictionaries contain patches with
magnetogram gradients (see Figs. 4 and 5 in Moon et al~\cite{moon2015partII}),
the distributions of the corresponding coefficients within the neutral
line may be useful for distinguishing between complex and simple ARs
and thus lead to the decreased bounds on the BER and improved classification.

The NMF results are not shown, but similar trends are observed. For
both the Grassmannian and Hellinger metrics, the estimated bounds
and the SVM error rate generally decrease as the weight increases,
suggesting that the neutral line is better suited for this classification
problem than the data from within the sunspots when using NMF dictionaries.
However, the estimated bounds, confidence intervals, and error rates
are generally still high (>0.25).

In general, these results indicate that if the goal is to accurately
classify SSs into complex or simple SSs based on the Mount Wilson
definition, then additional or different features are required. The
dictionary features may still be relevant for other learning tasks
such as predicting and detecting solar eruptive events.

\section{Conclusion}

\label{sec:conclusion}Applying meta learning or ensemble methods
to the problem of estimating $f$-divergence functionals results in
more accurate estimates. This ensemble estimator is useful for estimating
multiple bounds on the Bayes error rate. By simulation, we found that
the $\tilde{D}_{q_{1}}$ bounds are more accurate than the Chernoff
bound and the $G_{\alpha}$ bound in the sense that they are tighter
for all values of the BER. The $G_{\alpha}$ bound, however, is closer
to the BER when it is small and when the dimension is low. The MST
and weighted $k$-nn estimators had similar performance, suggesting
that the MST based method may converge rapidly to the true value in
at least some circumstances.

From the BER bounds of the sunspot data, we found that learned SVD
dictionaries from the neutral line are unlikely to be helpful in classifying
SSs (as either a simple SS or complex SS) based on the Mount Wilson
definition. However, including the dictionary coefficients from the
neutral line does seem to result in lower bounds on the BER and better
classification performance than when just using the dictionary coefficients
from within the sunspots. Overall, additional features are likely
necessary to achieve accurate classification of sunspots into these
categories.


\bibliographystyle{IEEEbib}
\bibliography{KevinSPW}

\end{document}